\documentclass[nopreprintline,12pt]{elsarticle}

\usepackage{amssymb}

\usepackage{graphicx}
\usepackage{algorithm}
\usepackage{algpseudocode}
\usepackage{algorithmicx}
\usepackage{amsmath}
\usepackage{subcaption}
\usepackage{multirow}
\begin{document}

\begin{frontmatter}

\title{Activator: GLU Activation Function as the Core Component of a Vision Transformer}

\author[inst1]{Abdullah Nazhat Abdullah }

\author[inst1]{Tarkan Aydin}

\affiliation[inst1]{organization={Bahcesehir University},
            city={Istanbul},
            country={Turkiye}}

\begin{abstract}
The transformer architecture has driven many successes in a variety of tasks within the field of deep learning, in particular the recent advances in natural language processing (NLP) culminating with large language models (LLM). Adding to that success, transformer architecture has found widespread interest from computer vision (CV) researchers and practitioners, allowing for many advancements in vision-related tasks and opening the door for multi-task and multi-modal deep learning architectures that share the same principle of operation. One drawback to these architectures is their reliance on the scaled dot product attention mechanism with the softmax activation function, which is computationally expensive and requires large compute capabilities for both training and inference. This paper investigates substituting the MLP and attention mechanism usually adopted for transformer architecture with an architecture based on incorporating a gated linear unit (GLU) activation function structure with the aim of reducing the computational cost. The equalized experimental assessments conducted in this work show that the proposed modification with the targeted reductions in computational complexity offers competitive performance compared to the selected baseline architectures. The results are significantly in support of the aims of this work, in which the focus was to extensively utilize GLU-based MLPs, establishing a more efficient but capable alternative to the traditional MLP and the attention mechanism as the core component in the design of transformer architectures.

\end{abstract}


\end{frontmatter}

\section{Introduction}

\noindent The transformer architecture \cite{1} has been established as a highly generalized and effective deep learning design. With the attention mechanism at the core of transformers, many advancements have been made in natural language processing (NLP), which culminated with the introduction of large language models (LLM) \cite{2} such as GPT \cite{3},LLama \cite{4}, Falcon \cite{5}, and Mistral \cite{6}. This interest also continued to the computer vision (CV) domain, resulting in several implementations, including ViT \cite{7}, MLP-Mixer \cite{8}, Conv-Mixer \cite{9}, Swin Transformer \cite{10}, Detection Transformer (DETR) \cite{11}, Perceiver-IO \cite{12}, Unified-IO \cite{13}, DINO \cite{14}, and Segment Anything Model (SAM) \cite{15}. This expanded adoption of the transformer architecture led to the inquiry of formulating lightweight and less computationally demanding variants, with examples of such attempts including Linformer \cite{16}, FNets \cite{17}, Local-ViT \cite{18}, Max-ViT \cite{19}, and Nystromformer \cite{20}. Many computer vision tasks require stringent computational resources, and the traditional transformer architecture with its global token-to-token interaction through the scaled dot product attention with the softmax activation function \cite{1} demands heavy allocation of compute capacity that is not suitable for a variety of applied computer vision settings, and the attempts in the literature for lightweight modifications aim to replace the attention mechanism with either restricted forms of the attention mechanism or approximate alternatives of it that are still beyond the capacity of many edge compute devices, therefore limiting the general adoption of transformer-based designs. This work proposes a new transformer architecture with local computations based on the gated linear unit activation process \cite{21}, which opens the possibility of adoption on low-compute-capacity platforms. The proposed design aims to reduce the general structure present in a typical transformer represented by the token-to-token interaction process followed by a token-wise transform process. Our proposed architecture relies solely on the GLU MLPs as a single token-to-token interaction and token-wise transform, significantly reducing the compute requirement. The architecture introduced by this work is trained and comparatively evaluated with respect to multiple baselines representing a variety of architectural design principles, all within an equalized setting to ensure a fair comparison. The experimental evaluations obtained showed that the newly introduced GLU activation-based transformer architecture (Activator) performed competitively compared to baseline architectures, validating the capability of the selected core functions of GLU-based MLPs as the basis structure of this proposed architecture.\noindent The main contributions of our work are the following: \begin{itemize}\item A novel computational block that replaces the softmax attention and the default MLP with GLU activation function-based MLPs.\item An extensive experimental validation of using only one layer of GLU MLPs as the sole component of the transformer computational block with ablation of different activation variants of GLU MLPs.\item Investigate the capability of local token interactions in contrast to the computationally heavy global attention mechanism. \end{itemize}

\section{Related Work}

\noindent   A review of the literature reveals a significant interest in investigating and testing modifications to the standard transformer architecture \cite{22},\cite{23},\cite{24},\cite{25}. Beltagy et al. introduced Longformer\cite{26}, which also uses a combination of band attention and internal global-node attention. Classification tokens are selected as global nodes. The architecture substitutes the band attention heads in the upper layers with dilated window attention, thus increasing the receptive field without increasing computation. Guo et al. introduced Star Transformer \cite{27}, combining band attention and global attention. This formulation of the transformer has a global node on which a band attention of width 3 is applied. Also, a shared global node connects a pair of non-adjacent nodes, while adjacent nodes are connected to each other. Katharopoulos et al. proposed the Linear Transformer \cite{28} with feature maps that target an approximation of the full softmax-activated scaled dot product attention and showed comparable performance in empirical tests. Wang et al. introduced Linformer \cite{16}, showing an approximation to the attention mechanism by a low-rank matrix, thus lowering the computational requirement while maintaining comparable performance. Choromanski et al. proposed Performer \cite{29}, which uses random feature maps as an approximation to the traditional attention function. Wang et al. introduced the Cascade Transformer \cite{30} By using a sliding window attention, the window size is exponentially increased when increasing the number of layers, leading to a reduction in complexity. Li et al. introduced the LogSparse Transformer \cite{31} that facilitates long-term dependency on time series analysis by using Eponym attention. Qiu et al. introduced BlockBERT \cite{32}, which uses block-wise attention to split the input sequence into non-overlapping blocks. Tay et al. introduced the sparse Sinkhorn attention \cite{33}. This mechanism is essentially block-wise attention, but the keys are sorted block-wise, therefore learning the permutations. Dai et al. proposed the Transformer-XL\cite{34}. This design uses a recurrence between the windows that is segment-based by storing the representations of the previous window and storing them in first-in, first-out memory (FIFO). After this step, the Transformer-XL applies attention to the sorted representations that have been stored in memory. Kitaev et al. introduced Reformer \cite{35} as a modified transformer that employs locality-sensitive hashing (LSH). The LSH is used to select the key and value pairs for each query, therefore allowing each token to attend to tokens that exist in the same hashing bucket. BigBird architecture by Zaheer et al.\cite{36} utilizes random attention to approximate full attention with a sparse encoder and sparse decoder, and it was shown by the analysis that this design can simulate any Turing Machine, explaining the capability of such architecture. Clustered Attention, proposed by Vyas et al. \cite{37} clusters the queries and then calculates the attention distributions for cluster centroids. Zhang et al. proposed PoolingFormer \cite{38}, which utilizes a two-level attention, a sliding window attention, and a compressed memory attention. The compressed memory module is used after first applying the sliding window attention and then applying a compressed memory module for the purpose of increasing the receptive field. Liu et al. proposed Memory Compressed Attention (MCA) \cite{39}, which complements local attention with strided convolution, thus reducing the number of keys and values. This allows the architecture to process much longer sequences compared to traditional transformers. Xiong et al. used the Nyström method to modify Transformer with the introduction of Nyströmformer \cite{20}. This design selects landmark nodes by the process of strided average pooling and then processes these selected queries and keys with an approximation to attention by the Nyström method. Funnel Transformer \cite{40} was proposed by Dai et al. by employing a funnel-like encoder that has a gradual reduction of the hidden sequence length using pooling along the sequence dimension; the proper length is then restored with an up-sampling process. gMLP \cite{41} was introduced by Liu et al., and this architecture is comprised of a series of blocks that are homogeneous in size and width. Each block layout is highly reminiscent of inverted bottlenecks. Another feature of this architecture compared to traditional transformers is that it does not require position embeddings. Local-ViT \cite{18} was introduced by Li et al. This architecture incorporates 2D depth-wise convolutions instead of the feed-forward network as in ViT. This design choice was inspired by the inverted residuals of MobileNets. Max-ViT \cite{19} was introduced by Tu et al., which repeats the basic building block over multiple stages. The basic block consists of two aspects: blocked local attention and dilated global attention. Ho et al. proposed the Axial Transformer \cite{42}. This architecture computes a sequence of attention functions, with each one applied along a single axis of the input, reducing the computational cost. Swin Transformer \cite{10} is an architecture proposed by Liu et al., and this design reduced the cost by splitting the image input into non-overlapping patches. These patches are then embedded as tokens for processing by attention. FNets \cite{17} was introduced by Lee-Thorp et al., and it proposes an attention-free transformer architecture that substitutes the scaled dot product attention with the softmax activation function. The Fourier sublayer applies a 2D DFT to the embedded input in two steps: one 1D DFT along the sequence dimension and another 1D DFT along the hidden dimension. Synthesizer \cite{43} was proposed by Tay et al. as an architecture that learns synthetic attention weights and does not rely on interactions between tokens. The results showed competitive performance in relation to other linear transformer designs. Transformer iN Transformer (TNT) \cite{44} was introduced by Han et al. This design treats the input images in a similar manner to a paragraph of text and divides them into several patches as “visual sentences” and then further divides them into sub-patches as “visual words." With this hierarchical division, the architecture is divided into conventional transformer blocks for extracting features and attentions on the visual sentence level, and then a sub-transformer is introduced in order to extract the features of smaller visual words. De et al. proposed Hawk and Griffin models \cite{45}; these are hybrid models combining gated linear recurrences and local attention with good extrapolation capabilities. MABViT was proposed by Ramesh and Ramkumar \cite{46} and introduced a transformer variant by integrating GLU non-linearity within the attention block, providing improved performance when run with the attention in parallel for the image classification task.

\newpage
\section{Methodology}
\noindent The methodology section is divided into two subsections. In the first subsection, the baseline architectures used in the evaluation are outlined, followed by a second subsection where our proposed Activator architecture is described.

\subsection{Baselines}	
\noindent For an extensive comparative analysis of capability, our proposed architecture is contrasted to two baseline architectures that represent distinct functional principles. The ViT follows the principles of a traditional NLP transformer, which represented the first iteration of designs that adopted such architecture. At its core, it relies on the scaled dot product attention with the softmax activation function, and in similarity to NLP-oriented transformers, the Vit also introduced the homogeneous layer structure.

\noindent Equations (1), (2) and (3) are the main equations for the ViT block.

\begin{equation}
\text{Attention(Q,K,V)} =\text{softmax}\left(\dfrac{QK^T}{\sqrt{d_k}}\right) V 
\end{equation}

\begin{equation}
\text{Y(X)} = 
\text{Attention}(\text{LayerNorm(X)})+ \text{X}
\end{equation}

\begin{equation}
\text{Z(Y)} = 
\text{MLP}(\text{LayerNorm(Y)})+\text{Y} 
\end{equation}

\newpage
\noindent Procedure 1 overviews the ViT block.

\begin{algorithm}[H]
\caption{\textbf{Procedure 1 : }ViT}
\begin{algorithmic}

\State{\textbf{Input:}} Image $I$, number of classes $C$, patch size $ps$, embedding dimension $d_{model}$, number of Transformer blocks $B$, hidden dimension of MLP $d_{mlp}$, learning rate $\eta$

\State{\textbf{Output:}} Predicted class probabilities

\State{\textbf{Steps:}}

\hspace*{5mm}1. Divide $I$ into patches of size $ps \times ps$.

\hspace*{5mm}2. Flatten each patch and embed it into a $d_{model}$-dimensional vector \hspace*{18mm}using patch embedding layer.

\hspace*{5mm}3. Concatenate the embedded patches into a sequence $X$.

\hspace*{5mm}4. \textbf{for} $i = 1$ to $B$ \textbf{do:}

    \hspace*{10mm}* Branch $X$ into residual and nonresidual paths.
    
    \hspace*{10mm}* Normalize the nonresidual path and Apply Attention.
    
    \hspace*{10mm}* Add the residual path.
    
    \hspace*{10mm}* Branch Attention result into residual and nonresidual paths.
    
   \hspace*{10mm}* Normalize the nonresidual path and Apply MLP block.
   
    \hspace*{10mm}* Add the residual path.
    
\hspace*{5mm}5.\hspace*{3mm}Apply global average pooling to the output of the last 
\hspace*{20mm}Transformer block.

\hspace*{5mm}6. Use a fully connected layer with $C$ output units and softmax \hspace*{20mm}activation to obtain class probabilities.

\hspace*{5mm}7. Train the model by minimizing the loss between predicted and \hspace*{20mm}true labels using gradient descent with learning rate $\eta$.

\end{algorithmic}
\end{algorithm}

\noindent The MLP-Mixer adopts the homogeneous layer structure as with the ViT but introduces efficiency-oriented computational operations of mixing (interacting) the token representation with the application of MLP that are applied in two successive stages: first, an MLP mixing of per token representation, and second, a per position (channel) MLP mixing of representations in between the tokens.

\noindent Equations (4) and (5) are the main equation for the MLP-Mixer block.

\begin{equation}
\text{Y(X)} = \text{Transpose}(\text{MLP}(\text{Transpose}(\text{LayerNorm(X)})))
+\text{X} 
\end{equation}

\begin{equation}
\text{Z(Y)} = 
\text{MLP}(\text{LayerNorm(Y)})+\text{Y} 
\end{equation}

\noindent Procedure 2 overviews the MLP-Mixer block.

\begin{algorithm}[H]
\caption{\textbf{Procedure 2 : }MLP-Mixer}

\begin{algorithmic}

\State{\textbf{Input:}} Image $I$, number of classes $C$, patch size $ps$, embedding dimension $d_{model}$, number of Transformer blocks $B$, hidden dimension of MLP $d_{mlp}$, learning rate $\eta$

\State{\textbf{Output:}} Predicted class probabilities

\State{\textbf{Steps:}}

\hspace*{5mm}1. Divide $I$ into patches of size $ps \times ps$.

\hspace*{5mm}2. Flatten each patch and embed it into a $d_{model}$-dimensional vector \hspace*{20mm}using patch embedding layer.

\hspace*{5mm}3. Concatenate the embedded patches into a sequence $X$.

\hspace*{5mm}4. \textbf{for} $i = 1$ to $B$ \textbf{do:}

    \hspace*{10mm}* Branch $X$ into residual and nonresidual paths.
    
    \hspace*{10mm}* Normalize the nonresidual path and Transpose.
    
    \hspace*{10mm}* Apply MLP block.
    
    \hspace*{10mm}* Transpose.
    
    \hspace*{10mm}* Add the residual path.
    
    \hspace*{10mm}* Branch result into residual and nonresidual paths.
    
   \hspace*{10mm}* Normalize the nonresidual path and Apply MLP block.
   
    \hspace*{10mm}* Add the residual path.
    
\hspace*{5mm}5.\hspace*{3mm}Apply global average pooling to the output of the last 
\hspace*{20mm}Transformer block.

\hspace*{5mm}6. Use a fully connected layer with $C$ output units and softmax \hspace*{20mm}activation to obtain class probabilities.

\hspace*{5mm}7. Train the model by minimizing the loss between predicted and \hspace*{20mm}true labels using gradient descent with learning rate $\eta$.

\end{algorithmic}
\end{algorithm}

\newpage
\subsection{Proposed Architecture}
\noindent This work investigates the utilization of gated liner unit-based transformer blocks, with token local operations being the core process of the architecture design. To achieve this aim, we propose replacing both the attention mechanism and the standard MLP with a single GLU MLP structure. The GLU MLP incorporates two up projections of the input token from the token dimension value to the expanded hidden dimension value; one of the paths is used as input to an activation function such as the gaussian error linear unit activation function (GELU), and the resulting structure has similarities to structures based on the GLU such as the gaussian error gated linear unit MLP (GEGLU) \cite{47}. We modify the general GLU MLP design by omitting the reduction of the hidden dimension by the 2/3 factor, and this proposal maintains the full hidden dimension to provide a suitable learning capacity to operate as a replacement for the attention mechanism. The second path is not subjected to non-linearity, and the two paths are then coupled by a process of element-wise multiplication that induces a gating on the non-activated path by the output of the activation. Therefore, the use of local token interactions as the core process leads to a reduction of computational cost.

\noindent This work adopts the Activator name for the proposed design to emphasize the GLU as the core process in designing the computational block.

\noindent Equations (6), (7) and (8) are the main equations for the Activator block.

\begin{equation}
    \text{P1(X)} = \text{LayerNorm(LinearUpProject(LayerNorm(X))}
\end{equation}

\begin{equation}
    \text{P2(X)} = \text{LayerNorm(GELU(LinearUpProject(LayerNorm(X)))}
\end{equation}

\begin{equation}
\text{Y(P1,P2)} = \text{LinearDownProject(P1 * P2)}
+\text{X} 
\end{equation}

\newpage
\noindent Procedure 3 overviews the Activator block.

\begin{algorithm}[H]
\caption{\textbf{Procedure 3 : }Activator}

\begin{algorithmic}

\State{\textbf{Input:}} Image $I$, number of classes $C$, patch size $ps$, embedding dimension $d_{model}$, number of Transformer blocks $B$, hidden dimension of MLP $d_{mlp}$, learning rate $\eta$

\State{\textbf{Output:}} Predicted class probabilities

\State{\textbf{Steps:}}

\hspace*{5mm}1. Divide $I$ into patches of size $ps \times ps$.

\hspace*{5mm}2. Flatten each patch and embed it into a $d_{model}$-dimensional vector \hspace*{20mm}using patch embedding layer.

\hspace*{5mm}3. Concatenate the embedded patches into a sequence $X$.

\hspace*{5mm}4. \textbf{for} $i = 1$ to $B$ \textbf{do:}

    \hspace*{10mm}* Branch $X$ into residual and nonresidual paths.
    
    \hspace*{10mm}* Normalize the nonresidual path and Branch into two projection   \hspace*{22mm}streams.
    
    \hspace*{10mm}*\hspace*{2mm}Apply a Linear Up Projection to the value of the 
    \hspace*{22mm}hidden dimension on each of the two streams.
    
    \hspace*{10mm}* Apply a GELU non linearity on only one of the streams and \hspace*{22mm}element-wise multiply the streams. 
    
    \hspace*{10mm}*\hspace*{2mm}Apply a Linear Down Project to the value of the token \hspace*{22mm}dimension and add the residual path.

\hspace*{5mm}5.\hspace*{3mm}Apply global average pooling to the output of the last 
\hspace*{20mm}Transformer block.

\hspace*{5mm}6. Use a fully connected layer with $C$ output units and softmax \hspace*{20mm}activation to obtain class probabilities.

\hspace*{5mm}7. Train the model by minimizing the loss between predicted and \hspace*{20mm}true labels using gradient descent with learning rate $\eta$.

\end{algorithmic}
\end{algorithm}

\begin{figure}
  \centering
  
   \includegraphics[width=4.0in,height=4.5in]{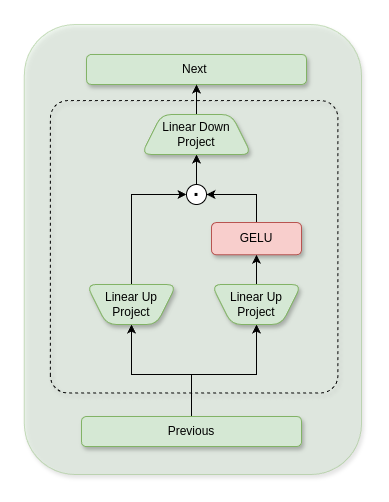} \\[\abovecaptionskip]

  \caption{An illustration of the Activator mechanism.}\label{Blocks}
\end{figure}

\begin{figure}[H]
    \centering
    
    ~ 
    \begin{subfigure}[t]{0.5\textwidth}
        \centering
        \includegraphics[width=\textwidth,height=3.0in]{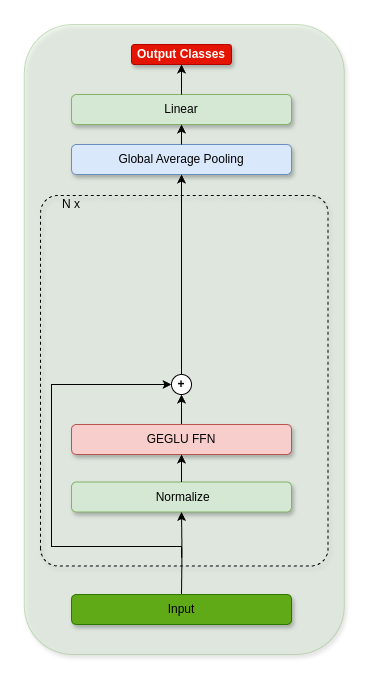}
        \caption{Activator architecture}
    \end{subfigure}
      ~ 
    \begin{subfigure}[t]{0.5\textwidth}
        \centering
        \includegraphics[width=\textwidth,height=3.0in]{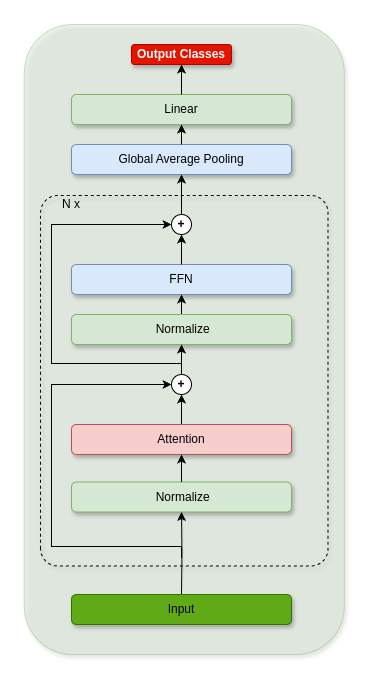}
        \caption{ViT architecture}
    \end{subfigure}
   
    \caption{A diagrammatic comparison of the Activator architectures with the ViT architecture.}
    \label{architectures}
\end{figure}

\noindent Fig. \ref{Blocks} shows the Activator mechanism, while Fig. \ref{architectures} compares the proposed architectures with that of ViT.

\newpage
\section{Results}

\noindent In this work, the experimental process utilized two data sets, as follows:
\begin{enumerate} 
\item The CIFAR-10 \cite{48} dataset consists of 60000 color images in 32 by 32 resolution provided for 10 classes, with 6000 images per class. There are 50000 training images and 10000 test images. \item The CIFAR-100 \cite{48}dataset consists of 60000 color images in 32 by 32 resolution; the number of classes is 100, resulting in 600 images per class. Similar to CIFAR-10, there are 50000 training images and 10000 test images.\end{enumerate}
The utilized software tools are as follows:
\begin{enumerate}
\item Python programming language of version 3.9. \item Pytorch framework of version 1.13. 
\item NVIDIA CUDA toolkit, of version 11.6.2.\end{enumerate}
The available hardware system is specified as the following:\begin{enumerate} 
\item Intel i9-9900k CPU.
\item 32 Gigabytes of system RAM. 
\item Nvidia RTX 2080ti GPU with 12 Gigabytes of VRAM. 
\item UBUNTU 20 LTS operating system.\end{enumerate}
The implementation details of the selected transformer architectures in this work are as follows:
\begin{enumerate} 
\item For the ViT architecture, the chosen patch size was 4 with a token dimension of 256, and the number of layers chosen was 4 with 4 attention heads and an MLP dimension of 512. 
\item For the MLP-Mixer architecture, the chosen patch size was 4 with a token dimension of 256, and the number of layers chosen was 4 with a token-wise MLP dimension of 512 and a channel-wise MLP dimension of 512.

\item For Activator, the chosen patch size was 4, with a token dimension of 256 and an expansion dimension of 512. The number of layers chosen was 4.
\end{enumerate}

\noindent All models were fitted with a training loop comprised of 100 epochs with a batch size of 128. All experiments adopted the recommended learning rate for the Adam optimizer of 0.001 \cite{49}.

\hfill

\noindent Table 1 illustrates the obtained results after performing the experimentation on CIFAR-10 and CIFAR-100 dataset applied to the baseline architectures and Activator architecture.

\begin{table}[htbp]
\caption{Experimental test accuracy in percentages (\%) obtained on the utilized dataset.}
\begin{tabular}{|l|rr|}
\hline
\multirow{2}{*}{Models} & \multicolumn{2}{l|}{Data sets}                                                                            \\ \cline{2-3} 
                                & \multicolumn{1}{p{47mm}|}{CIFAR-10}        & \multicolumn{1}{p{47mm}|}{CIFAR-100} \\ \hline
ViT                               & \multicolumn{1}{r|}{65.74}          & 34.87                         \\ \hline
MlpMixer                      & \multicolumn{1}{r|}{70.12}          & 39.16                         \\ \hline

Activator \textbf{(ours)}                & \multicolumn{1}{r|}{\textbf{73.2}} & \textbf{46.14}                \\ \hline
\end{tabular}
\end{table}

\begin{figure}[H]
    \centering
    \begin{subfigure}[t]{0.5\textwidth}
        \centering
        \includegraphics[width=\textwidth,height=2.7in]{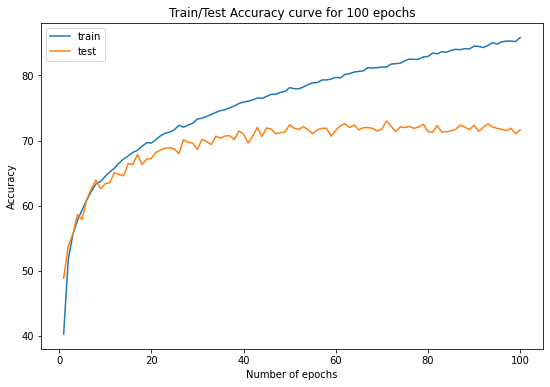}
        \caption{CIFAR-10 accuracy curve}
    \end{subfigure}%
    ~ 
    \begin{subfigure}[t]{0.5\textwidth}
        \centering
        \includegraphics[width=\textwidth,height=2.7in]{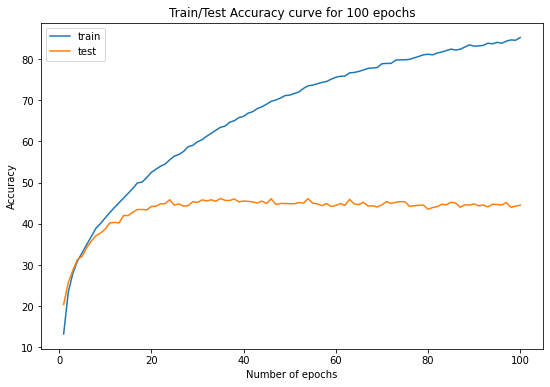}
        \caption{CIFAR-100 accuracy curve}
    \end{subfigure}
    
    \caption{An illustration of the accuracy curves for Activator architecture.}

    \label{accuracy2}
\end{figure}

\noindent Fig.\ref{accuracy2} shows the accuracy curves obtained on Activator architecture for the CIFAR-10, CIFAR-100 datasets.

\begin{figure}[H]
    \centering
    \begin{subfigure}[t]{0.5\textwidth}
        \centering
        \includegraphics[width=\textwidth,height=2.7in]{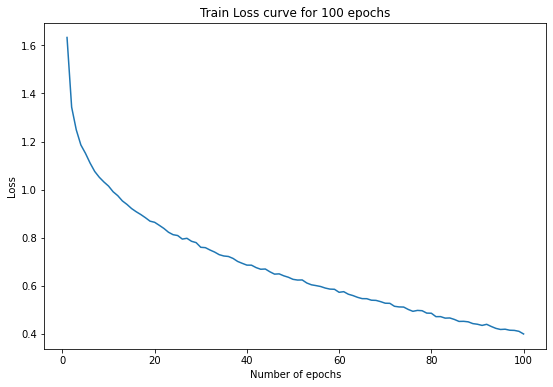}
        \caption{CIFAR-10 loss curve}
    \end{subfigure}%
    ~ 
    \begin{subfigure}[t]{0.5\textwidth}
        \centering
        \includegraphics[width=\textwidth,height=2.7in]{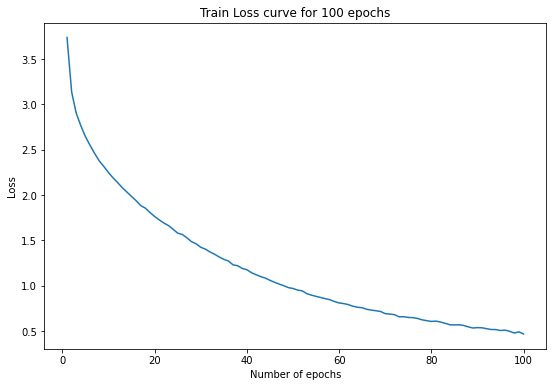}
        \caption{CIFAR-100 loss curve}
    \end{subfigure}

    \caption{An illustration of the loss curves for Activator architecture.}

    \label{loss2}
    
\end{figure}

\newpage
\noindent Fig.\ref{loss2} shows the loss curves obtained on Activator architecture for the CIFAR-10, CIFAR-100 datasets.

\section{Ablation Studies}

\noindent To further validate the robustness of the proposed mechanism and its capability, the type of the activation function used in the GLU MLP is ablated to include, in addition to the gaussian error linear unit activation function that results in a (GEGLU) Activator mechanism, the rectified linear unit GLU MLP (ReGLU) \cite{47} Activator and the Swish GLU MLP (SwiGLU) \cite{47} Activator.

\noindent Table 2 provides a comparison of the performance between the ablated Activator variants.

\begin{table}[htbp]
\caption{Ablation test accuracy in percentages (\%) obtained on the utilized dataset.}
\begin{tabular}{|l|rr|}
\hline
\multirow{2}{*}{Models} & \multicolumn{2}{l|}{Data sets}                                                                            \\ \cline{2-3} 
                                & \multicolumn{1}{p{47mm}|}{CIFAR-10}        & \multicolumn{1}{p{47mm}|}{CIFAR-100} \\ \hline
Activator\_GEGLU                               & \multicolumn{1}{r|}{73.2}   & 46.14                          \\ \hline
Activator\_SwiGLU                      & \multicolumn{1}{r|}{72.85}   & 46.07                        \\ \hline
Activator\_ReGLU                        & \multicolumn{1}{r|}{72.95}   & 46.12                        \\ \hline

\end{tabular}
\end{table}

\noindent The ablation results show no strong deviation or drop in the performance metrics of our proposed architecture by the introduction of variation of the activation function. This finding adds validation that the Activator mechanism is a suitable candidate for a core component in constructing Transformer models.

\section{Conclusion}

\noindent This work proposed a transformer architecture modification by adopting GLU-based MLPs as a replacement for the traditional MLP and the attention mechanism, utilizing the modified GLU-based MLPs as the sole process in the design of the core block in transformer architectures. The learned gating function, with its element-wise multiplication and linear weight projections, offers a significantly more lightweight and much simpler mechanism to compute the dynamic information filtering that traditional attention is known to implement. Additionally, using GLU-based MLPs alone can also provide the learned key-value mapping capability that the feed-forward MLP of the traditional transformer offers separately from attention; thus, a GLU-based MLP adopted for this investigation presents a more cohesive computational unit for constructing transformers. The proposals of this work are strongly supported by the experimental results, which show that GLU MLPs as the only core process can compete with and outperform the baseline architectures, opening the door to utilizing the proposed design in the multitude of tasks and modalities for which attention-based transformers were previously the forerunner generalist deep learning central component.

\newpage

\begin{itemize}

\item \textbf{Author's emails}

Abdullah Nazhat Abdullah: nazhat.abdullah@bahcesehir.edu.tr

Tarkan Aydin: tarkan.aydin@bau.edu.tr

\item \textbf{Author's ORCID IDs}

Abdullah Nazhat Abdullah: 0000-0002-1757-0785

Tarkan Aydin: 0000-0002-2018-405X
\end{itemize}

\end{document}